\documentclass{article} 
\usepackage{nips14submit_e,times}
\usepackage{hyperref}
\usepackage{url}
\usepackage{amssymb}
\usepackage[english]{babel}
\usepackage{graphicx}
\usepackage[utf8]{inputenc}
\usepackage{csquotes}
\usepackage{amsmath}
\usepackage{caption,subcaption}

\usepackage[]{algorithm2e}

\title{Activity report analysis with automatic single or multi span answer extraction}

\author{Ravi Choudhary , Arvind Krishna Sridhar, Erik Visser \\
	Qualcomm QTI Technologies Inc\\
	ravichou@qti.qualcomm.com, arvisrid@qti.qualcomm.com, evisser@qti.qualcomm.com
}

\nipsfinalcopy 
\providecommand{\keywords}[1]
{
	\small
	\textbf{\textit{Keywords---}} #1
}

\begin{document}
\maketitle
\begin{abstract}
In the era of loT (Internet of Things) we are surrounded by a plethora of Al enabled devices that can transcribe images, video, audio, and sensors signals into text descriptions. When such transcriptions are captured in activity reports for monitoring, life logging and anomaly detection applications, a user would typically request a summary or ask targeted questions about certain sections of the report they are interested in. Depending on the context and the type of question asked, a question answering (QA) system would need to automatically determine whether the answer covers single-span or multi-span text components. Currently available QA datasets primarily focus on single span responses only (such as SQuAD[4]) or contain a low proportion of examples with multiple span answers (such as DROP[3]). To investigate automatic selection of single/multi-span answers in the use case described, we created a new smart home environment dataset comprised of questions paired with single-span or multi-span answers depending on the question and context queried. In addition, we propose a RoBERTa[6]-based multiple span extraction question answering (MSEQA) model returning the appropriate answer span for a given question. Our experiments show that the proposed model outperforms state-of-the-art QA models on our dataset while providing comparable performance on published individual single/multi-span task datasets.  
\end{abstract}


\section{Introduction}

Machine reading comprehension[1], specifically extractive closed-domain question-answering, is an active field of research in natural language processing. Solving question-answering tasks has a variety of real-world applications, in particular in Al assistants for personal and business use (clinical[2], retail distribution[3] etc). Extractive single span question- answering has recently advanced to the point where machines can now compete with, if not outperform, human question-answering capability[4]. 
    
 Progress in question answering can be tracked by reviewing performance on typical bench- mark databases and tasks[5,6,8,9]. Most state-of-the-art question-answering datasets consist of only single-span question-answer pairs, or answers consisting of only one single extracted word. The SQUAD[5] dataset, which has over 100,000 single-span questions and is larger than most previous question-answering datasets, is one example of this type of dataset. SQUAD has several answer types, including Date, Person, and Location, and its passages cover a wide range of topics. The SQUAD 2.0[5], combines the original ques- tions in with new unanswerable questions, forcing models to learn when to refrain from answering. Likewise, the DROP[6,7] datasets are also extractive, closed-domain datasets with a low proportion of multi-span question-answer pairs (6.0 percent for DROP and 15.68 percent for MASHQA[8]). Compared to SQUAD, the questions in DROP are more involved to force the system to have a deeper understanding of the passages semantics and answer them via discrete reasoning. HotpotQA[9] dataset focuses on multi-hop questions, which require accessing multiple documents to generate an answer.
 
 Some of the most popular QA models are based on BERT[4] and its variations, RoBERTa[10], MobileBERT[11] and ALBERT[12]. BERT[4] is capable of executing a variety of general NLP tasks, such as next sentence prediction, masked Language Modeling (LM), Natu- ral Language Inference (NLI), etc . RoBERTa, or Robustly Optimized BERT approach, introduces alternative strategies for the BERT training process to improve performance. Key differences between RoBERTa and BERT include longer training times and an order of magnitude more training data. A BERT based multi span QA model MTMSN([13] predicts a number of spans as a classification task and then picks top K spans from a single QA model. From these top spans, it then picks non- overlapping spans using non-maximum suppression (NMS) algorithm[14] by removing unnecessary tokens from the answer span. Tag-based-multispan prediction(TASEIO + SSE)[15] is based on a ROBERTa model and solves a task similar to Named Entity Tags (NER) or Part of Speech (POS) tasks. Finally MASHQA[8] model, built on Transformer- XL[16], classifies every sentence whether a sentence has an answer or not, by combining a representation of each sentence with a representation of paragraph and question.
 
 While these QA datasets and models address either single or multi span answering tasks, they do not address the problem of determining whether a single or multi span answer is needed for a particular question. This discrimination task is for example necessary in a scenario where users would like to query activity reports obtained from Al enabled devices generating text descriptions from various logged multimedia and sensors content. Such reports are generated in lifelogging, monitoring or anomaly detection applications over extended periods of time and may contain expected or unexpected events. This in turn may prompt a user to ask targeted questions about a known occurrence or more vague questions covering a longer period of time or range of topics. A user may for instance ask questions such as Where was X at 4pm? or What happened at location Y in the morning?. While the former requires a single span answer (i.e. Xs location), the latter question requires the system to gather a multi span answer from information spread out over a number of sentences (i.e. list all events that happened at location Y in the am). Figure 1 illustrates the various types of spans recovered depending on the question asked in the case of querying an activity report in a smart home use case.
 
 In this paper, we propose a ROBERTa[10] based QA system that can return the appropri- ate single or multi span answers depending on the question asked and the context being queried. Its performance is illustrated on a custom dataset for a smart home use case where multiple Al modules are transcribing image, video, audio and sensors signals into text descriptions and logging them in activity reports. We also show that the proposed QA system provides competitive performance on existing single and multi-span bench- mark datasets compared to state-of-the-art models. The paper is organized as follows: section 2 describes the custom dataset developed and proposed QA model architecture, section 3 the evaluation results and we conclude with section 4. 

  \begin{figure}    
   \centerline{\includegraphics[width=0.5\textwidth,clip=]{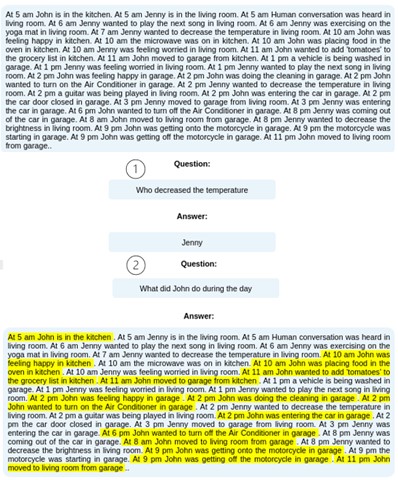}
              }
              \caption{Example of a (question, single-span answer) pair (1) and (question, multi-span answer) pair (2) for a given passage from a sample activity report: a single-span answer is appropriate for question (1) while question (2) requires a multi-span answer. A QA system for this task should be designed to return the appropriate answer span given the question and passage
                      }
   \label{F-simple}
   \end{figure}

\section{Approach}
\label{S-aug}
\subsection{Custom dataset generation process}

  To illustrate our proposed QA system, we evaluate its performance on prior art datasets as well as a custom in-house dataset for the smart home use case described. The latter, called SmartHome dataset, consists of paragraphs describing various user’s activity patterns in locations around a home throughout the day. It is logging information such as how a user was feeling, what commands they uttered when using smart assistants, activities detected from sound/sensor/video etc. in natural language form. We assume such natural language activity reports can be generated by aggregating descriptions from AI modules tagging video/camera/audio/sensor signal content with object detection, emotion detection, video/image/audio understanding etc. approaches. 
  
  To simulate the dataset passage, each generated context contains a list of sentences and each sentence is obtained by putting sensor/sound/video identified entities in a template along with a timestamp and location.  To simulate questions for such a dataset, we wrote a simple template based structural program to create a list of synthetic complex question starting with interrogative pronouns such as “what”, “when”, “where”, “who” as well as “Did he/she”, “Is/Was he”) for a given paragraph.  In addition, we included temporal events-based questions such as “When did X perform …”, “What did Y do before/after certain time”, “What emotional state Z was at a particular time”, etc. Figure 1 illustrates the content of the generated activity report and Figure 2 lists a few examples of typical questions. The type of question and their distribution is shown in Fig 2 (Visualization is created using a github repository https://github.com/mrzjy/sunburst).
  \begin{figure}[!h]
    \setlength\abovecaptionskip{0.2\baselineskip}
    \setlength\belowcaptionskip{-1.5\baselineskip}
    \centering
    \includegraphics[width=.9\textwidth,clip=]{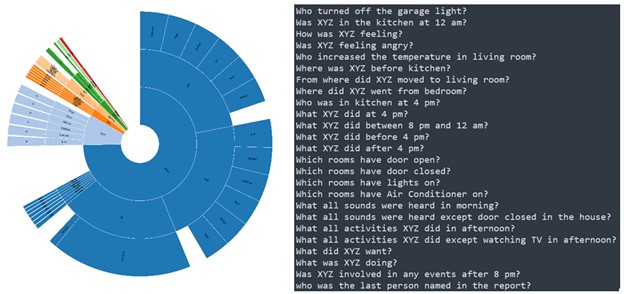}
    \caption{Trigram prefix distribution of the questions in SmartHome dataset and a list of some questions generated using a simple template based structural program.}
    \label{fig:nlp_fig2}
\end{figure}
   
   For a given sentence in a paragraph and created question, the corresponding answer text is obtained deterministically along with a start id and end id. Since sentences are generated using a template, the location of the answer is fixed and can be obtained by using some rules. For example, for the sentence “At 6 am, Jenny was exercising on the yoga mat in living room”, the index of time “6 am”, person name “Jenny”, activity description “exercising on the yoga mat” and location “living room” are known and fixed given the generation template described above. Hence, the answer to question “When Jenny was doing exercise?” can be extracted as “6 am” from index 1 of the generated sentence.
   
\subsection{Model Architecture}
   In this section, we describe each step from the pipeline of our proposed system called Multi Span Extraction Question Answering (MSEQA). The pipeline consists of a multi-task learning framework to select the appropriate answer span/type and is illustrated by Figure 3. Our model uses RoBERTa as encoder mapping word embeddings into contextualized representations using pre-trained Transformer blocks. Based on these representations, we employ a multi-type answer predictor providing the following classification: (1) span from the text; (2) answer Type; (3) multi-span/single span. We first predict the answer type of a given passage-question pair, and then adopt individual prediction strategies. To support multi-span extraction, the model explicitly predicts relevant sentences in a paragraph. 
   
\begin{figure}[!h]
    \setlength\abovecaptionskip{0.2\baselineskip}
    \centering
    \includegraphics[width=.9\textwidth,clip=]{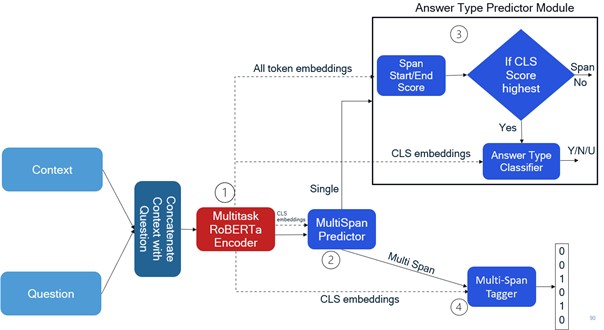}
    \caption{Proposed Multi-Span Extraction Question Answering (MSEQA) system overview}
    \label{fig:nlp_fig3}
\end{figure}

\subsubsection{Multitask RoBERTa Encoder}
   For a given token sequence X = [$x_{1}, x_{2},..., x_{T}$], RoBERTa[10], a deep Transformer network, outputs a sequence of contextualized token representations $H_{L}$ =[$h_{L1}, h_{L2},..., h_{LT}$]. RoBERTa-Base, denoted as block 1 in Fig 3, is made up of 12 Transformer layers (L = 12), each with 12 heads and a hL vector of size 768. Special markup tokens [CLS] and [SEP] are added to the beginning and end of the input sequence as an essential preprocessing step for RoBERTa. In cases where there are two separate input sequences, one for the question and one for the given context, such as MRC, an additional [SEP] is added between the two to form a single sequence.
\subsubsection{MultiSpan Predictor}
    The Multispan Predictor, denoted as block 2 in Figure 3, is a single FC connected layer which uses a hidden representation of [CLS] token to classify whether the question has a multi-span output or not. Execution is shifted to block 3 if the model predicts a single-span answer, and to block 4 if the model predicts a multi-span answer, as shown in Figure 3. It is done using categorial classification with
    
\begin{equation}  \label{Eq-H-def}
     p\_s = softmax(FFN([h_{L}[CLS]]))
\end{equation}
   
   where $h_{L}[cls]$ is the combined representation of question and context, p\_s is probability of question having single span answer or multi-span answer.
   
\subsubsection{Answer Type Predictor}
    The Answer Type prediction module, denoted as block 3, has two logically connected submodules, one identifying the single-span output and the other module determining if the answer is Yes, No or Unknown. Except for the Unknown answers, each answer is an extractive span in the passage or a Boolean answer. We use a FC layer to get the start logits and end logits for each token in the context. For Yes/No/Unk classification, a single FC layer is applied to the RoBERTa’s CLS output to obtain three logits for each class. We use start and end score of CLS against all other tokens to decide if we want to find a single span answer or logical reason answer (Yes\/ No\/ Unknown). If CLS has the highest score then we use a categorical variable to decide between the above three answer types, with probabilities, p\_a, computed as: 
    \begin{equation}
    p\_a = softmax(FFN([h_L[CLS]]))
    \end{equation}
    Otherwise, we find a span with highest score for a correct answer using the probabilities of the starting, p\_start, and ending positions, p\_end, from the passage as:
    \begin{equation}
    p\_start = softmax(FFN([H_L])
    \end{equation}
    \begin{equation}
    p\_end = softmax(FFN([H_L])
    \end{equation}
    where FFN is a two-layer feed-forward network with the GELU activation

\subsubsection{Multi-Span Tagger}
In our multispan tagger layer, denoted as block 4, we pair each sentence in the context with the question in the input format:  $<$CLS$><$QUESTION$<$SEP$>$SENTx where x is 1, 2,..n denote the sentence index. We classify the CLS representation to identify if the sentence is relevant for the given query as
\begin{equation}
p\_t = softmax(FFN([h_L[Q]:h_L[Senti]))
\end{equation}
where p\_t is probability of each sentence being an answer or not, h$_L$[Q] and h$_L$[Senti] is pooled contextualized representation of the question and ith sentence respectively.“:” denotes a concatenation operator.

\subsubsection{Objective}
The training objective of our multitask learning model is to minimize the following linear combination of question answering (L$_q$), answer-type prediction (L$_a$), multispan prediction (L$_s$) and multispan tagger losses (L$_t$), with $\lambda_q$, $\lambda_a$,  $\lambda_s$ and $\lambda_t$ being their respective weights.:
\begin{equation}
L = \lambda_qL_q + \lambda_aL_a + \lambda_sL_s + \lambda_tL_t
\end{equation}
Different tasks use different types of loss functions. The loss function for the question answering task i.e., L$_{q}$ is the loglikelihood of the correct start and end positions whereas the loss function for other tasks is cross entropy loss. Inspired by the work done in unifiedQA[17], we first pretrained the Roberta encoder on a combination of multiple datasets (SQuAD2.0, QuAC[18], HotPotQA[9], and BooleanQA[19]). The models in blocks 2, 3 and 4 on the other hand were randomly initialized and trained from scratch using multi-task objective (6).
\section{Experiments} 
      \label{S-general}      

\subsection{Datasets} 
  \label{S-text}
We evaluate our proposed model on three datasets, including SmartHome, DROP and MASHQA. 
\begin{enumerate}
\item SmartHome, as explained in 2.1, is made up of sequence of daily events which usually span 3 or more words. It comprises of 3.16k passages with 300k questions with a train-validation-test split of 80:12:8. 
\item DROP is a reading comprehension dataset comprising of complex questions that require Discrete Reasoning over Paragraphs. It’s passages were extracted from Wikipedia while the crowdsourced questions fall into three answer types – spans of text, date and numbers. Since our objective is to automate the selection of answer span types, we filtered out the QA pairs to contain only the spans of text as answer type as a part of our data preprocessing step. 
\item MASHQA is an extractive QA dataset, with multiple answer spans, curated by experts from consumer health domain. On an average, its passages are 3.3 times longer than DROP requiring the QA model to overcome long context problem.
\end{enumerate}

The latter two datasets are chosen to benchmark MSEQA on publicly available datasets and to represent different kinds of passage comprehension domains, passage \& answer lengths as well as reasoning types.
\begin{table}
\centering
\begin{tabular}{|lllllll|}
\hline
\multicolumn{1}{|c|}{Model}  & \multicolumn{2}{c|}{Single Span}                            & \multicolumn{2}{c|}{MultiSpan}                            & \multicolumn{2}{c|}{Overall}       \\ \hline
\multicolumn{1}{|l|}{}   & \multicolumn{1}{c}{EM} & \multicolumn{1}{c|}{F1} & \multicolumn{1}{c}{EM} & \multicolumn{1}{c|}{F1} & \multicolumn{1}{c}{EM} & F1 \\ \hline
\multicolumn{7}{|c|}{DROP}                                                                                                                                          \\ \hline
\multicolumn{1}{|l|}{MTMSN (BERT-base)} & \multicolumn{1}{c}{64.5}  & \multicolumn{1}{c|}{72.2}  & \multicolumn{1}{c}{23.1}  & \multicolumn{1}{c|}{64.8}  & \multicolumn{1}{c}{60.5}  & 71.5  \\ \hline
\multicolumn{1}{|l|}{MTMSN (BERT-large)}   & \multicolumn{1}{c}{70.5}   & \multicolumn{1}{c|}{78.8}   & \multicolumn{1}{c}{25.1}   & \multicolumn{1}{c|}{66.8}   & \multicolumn{1}{c}{65.8}   &  76.8  \\ \hline
\multicolumn{1}{|l|}{Tag-Based (Roberta-base)}   & \multicolumn{1}{c}{65.6}   & \multicolumn{1}{c|}{77.9}   & \multicolumn{1}{c}{37.3}   & \multicolumn{1}{c|}{65.1}   & \multicolumn{1}{c}{62.9}   &  76.5  \\ \hline
\multicolumn{1}{|l|}{Tag-Based (Roberta-large)}   & \multicolumn{1}{c}{70.6}   & \multicolumn{1}{c|}{78.4}   & \multicolumn{1}{c}{50.8}   & \multicolumn{1}{c|}{68.9}   & \multicolumn{1}{c}{68.5}   &  77.3  \\ \hline
\multicolumn{1}{|l|}{MSEQA (Roberta-Base)}   & \multicolumn{1}{c}{65.1}   & \multicolumn{1}{c|}{77}   & \multicolumn{1}{c}{46.9}   & \multicolumn{1}{c|}{67.9}   & \multicolumn{1}{c}{63.1}   &  76  \\ \hline
\multicolumn{1}{|l|}{MSEQA (Roberta-Large)} & \multicolumn{1}{c}{70.4}  & \multicolumn{1}{c|}{78.1}  & \multicolumn{1}{c}{50.6}  & \multicolumn{1}{c|}{68}  & \multicolumn{1}{c}{68.4}  & 77.2  \\ \hline
\multicolumn{7}{|c|}{SMARTHOME}                                                                                                                                          \\ \hline
\multicolumn{1}{|l|}{MTMSN (BERT -base)} & \multicolumn{1}{c}{69.5}  & \multicolumn{1}{c|}{79.6}  & \multicolumn{1}{c}{26.1}  & \multicolumn{1}{c|}{71.8}  & \multicolumn{1}{c}{53.7}  & 76.8  \\ \hline
\multicolumn{1}{|l|}{MTMSN (BERT -large)}   & \multicolumn{1}{c}{74.5}   & \multicolumn{1}{c|}{82.5}   & \multicolumn{1}{c}{31.1}   & \multicolumn{1}{c|}{72.8}   & \multicolumn{1}{c}{58.7}   &  79  \\ \hline
\multicolumn{1}{|l|}{Tag-Based (Roberta-base)}   & \multicolumn{1}{c}{69.7}   & \multicolumn{1}{c|}{79.9}   & \multicolumn{1}{c}{41.3}   & \multicolumn{1}{c|}{75.6}   & \multicolumn{1}{c}{59.4}   &  87.4  \\ \hline
\multicolumn{1}{|l|}{Tag-Based (Roberta-large)}   & \multicolumn{1}{c}{76.6}   & \multicolumn{1}{c|}{83.7}   & \multicolumn{1}{c}{59.8}   & \multicolumn{1}{c|}{75.9}   & \multicolumn{1}{c}{70.5}   &  80.9  \\ \hline
\multicolumn{1}{|l|}{MASHQA (XLNet)}   & \multicolumn{1}{c}{69.2}   & \multicolumn{1}{c|}{71.2}   & \multicolumn{1}{c}{63.2}   & \multicolumn{1}{c|}{70.1}   & \multicolumn{1}{c}{67}   &  70.8  \\ \hline
\multicolumn{1}{|l|}{MSEQA (Roberta-Base)}   & \multicolumn{1}{c}{67.9}   & \multicolumn{1}{c|}{82.3}   & \multicolumn{1}{c}{56}   & \multicolumn{1}{c|}{76}   & \multicolumn{1}{c}{63.6}   &  80  \\ \hline
\multicolumn{1}{|l|}{MSEQA (Roberta-Large)}   & \multicolumn{1}{c}{80}   & \multicolumn{1}{c|}{85.4}   & \multicolumn{1}{c}{64}   & \multicolumn{1}{c|}{82}   & \multicolumn{1}{c}{74.2}   &  84.2  \\ \hline
\multicolumn{7}{|c|}{MASHQA}                                                                                                                                          \\ \hline
\multicolumn{1}{|l|}{MASHQA (XLNet)}   & \multicolumn{1}{c}{23.2}   & \multicolumn{1}{c|}{70.9}   & \multicolumn{1}{c}{21.7}   & \multicolumn{1}{c|}{52}   & \multicolumn{1}{c}{21.55}   &  55  \\ \hline
\multicolumn{1}{|l|}{MSEQA (Roberta-Base)}   & \multicolumn{1}{c}{24.3}   & \multicolumn{1}{c|}{73.5}   & \multicolumn{1}{c}{21.8}   & \multicolumn{1}{c|}{52}   & \multicolumn{1}{c}{22.1}   &  55.3  \\ \hline
\multicolumn{1}{|l|}{MSEQA (Roberta-Large)}   & \multicolumn{1}{c}{24.6}   & \multicolumn{1}{c|}{74}   & \multicolumn{1}{c}{22.4}   & \multicolumn{1}{c|}{52.2}   & \multicolumn{1}{c}{22.7}   & 55.6   \\ \hline
\end{tabular}
\caption{The performance of the baseline models along with our multispan question answer model, MSEQA, trained on datasets individually}
\end{table}
\subsection{Implementation Details}
    Our model's implementation is based on the PyTorch[20] implementation of RoBERTa from a  huggingface[21] library. With a learning rate of 4e-5 and a batch size of 16, we use AdamW[22] as our optimizer. The epoch count is set to 10. Linear warmup strategy is employed for the first 6\% of steps, followed by a linear decay to 0. The maximum lengths for query and context sentences were fixed to 128 and 64 tokens, respectively. For the multispan tagger, we set number of sentences in a context (n) to 26.  The gradient norm is clipped to within one to avoid the gradient exploding problem. All texts are tokenized with BytePair Encoding (BPE)[23] and chopped into sequences of no more than 512 tokens. 

\subsection{Main Result}
To illustrate the effectiveness of MSEQA, we demonstrate its performance relative to state-of-the-art multi-span models reviewed in section 1 – TASEIO+SSE[15], MTMSN[13] and MASHQA[8]. In line with the models in comparison, we adopt F1 \& Exact Match (EM) scores as our evaluation metric. The F1 \& EM scores for DROP, MASHQA and SmartHome datasets are presented in Table 1 \& 2 for both base and large model types. We study the performance of the models on individual datasets in Table 1. It shows that MSEQA outperforms SOTA on SmartHome and MASHQA while it exhibits slightly worse results in DROP compared to TASEIO+SSE.
 
In Table 2, we trained each model with a combination of all the datasets but tested on individual dataset to see if the models can learn to generalize. From the Table 2 results, we can see a clear increase in F1 \& EM scores compared to Table 1 showing genuine improvement in the models’ multi span extraction abilities as it is notable that these datasets are from different domains. Furthermore, the relative performance trend of MSEQA is consistent with Table 1 showing that our proposed approach can improve with further pretraining on more complex reasoning datasets.

\begin{table}
\centering
\begin{tabular}{|lllllll|}
\hline
\multicolumn{1}{|c|}{Model}  & \multicolumn{2}{c|}{Single Span}                            & \multicolumn{2}{c|}{MultiSpan}                            & \multicolumn{2}{c|}{Overall}       \\ \hline
\multicolumn{1}{|l|}{}   & \multicolumn{1}{c}{EM} & \multicolumn{1}{c|}{F1} & \multicolumn{1}{c}{EM} & \multicolumn{1}{c|}{F1} & \multicolumn{1}{c}{EM} & F1 \\ \hline
\multicolumn{7}{|c|}{DROP}                                                                                                                                          \\ \hline
\multicolumn{1}{|l|}{MTMSN (BERT-base)} & \multicolumn{1}{c}{65.1}  & \multicolumn{1}{c|}{73.6}  & \multicolumn{1}{c}{23.6}  & \multicolumn{1}{c|}{65.4}  & \multicolumn{1}{c}{60.8}  & 72.8  \\ \hline
\multicolumn{1}{|l|}{MTMSN (BERT-large)}   & \multicolumn{1}{c}{71.2}   & \multicolumn{1}{c|}{79}   & \multicolumn{1}{c}{25.3}   & \multicolumn{1}{c|}{66.9}   & \multicolumn{1}{c}{66.5}   &  77.3  \\ \hline
\multicolumn{1}{|l|}{Tag-Based (Roberta-base)}   & \multicolumn{1}{c}{65.7}   & \multicolumn{1}{c|}{72.3}   & \multicolumn{1}{c}{37.5}   & \multicolumn{1}{c|}{65.2}   & \multicolumn{1}{c}{63.1}   &  72.9  \\ \hline
\multicolumn{1}{|l|}{Tag-Based (Roberta-large)}   & \multicolumn{1}{c}{71.8}   & \multicolumn{1}{c|}{79.4}   & \multicolumn{1}{c}{51.3}   & \multicolumn{1}{c|}{71.4}   & \multicolumn{1}{c}{69.7}   &  78.9  \\ \hline
\multicolumn{1}{|l|}{MSEQA (Roberta-Base)}   & \multicolumn{1}{c}{65.5}   & \multicolumn{1}{c|}{78.8}   & \multicolumn{1}{c}{48.9}   & \multicolumn{1}{c|}{68.9}   & \multicolumn{1}{c}{63.4}   &  78  \\ \hline
\multicolumn{1}{|l|}{MSEQA (Roberta-Large)} & \multicolumn{1}{c}{71}  & \multicolumn{1}{c|}{79.3}  & \multicolumn{1}{c}{51.6}  & \multicolumn{1}{c|}{72.6}  & \multicolumn{1}{c}{69}  & 77.2  \\ \hline
\multicolumn{7}{|c|}{SMARTHOME}                                                                                                                                          \\ \hline
\multicolumn{1}{|l|}{MTMSN (BERT -base)} & \multicolumn{1}{c}{70.2}  & \multicolumn{1}{c|}{79.8}  & \multicolumn{1}{c}{26.9}  & \multicolumn{1}{c|}{73.4}  & \multicolumn{1}{c}{54.5}  & 77.8  \\ \hline
\multicolumn{1}{|l|}{MTMSN (BERT -large)}   & \multicolumn{1}{c}{75.1}   & \multicolumn{1}{c|}{83.1}   & \multicolumn{1}{c}{31.9}   & \multicolumn{1}{c|}{74.9}   & \multicolumn{1}{c}{59.4}   &  79.8  \\ \hline
\multicolumn{1}{|l|}{Tag-Based (Roberta-base)}   & \multicolumn{1}{c}{70.3}   & \multicolumn{1}{c|}{80.3}   & \multicolumn{1}{c}{43.6}   & \multicolumn{1}{c|}{76.6}   & \multicolumn{1}{c}{60.6}   &  79.2  \\ \hline
\multicolumn{1}{|l|}{Tag-Based (Roberta-large)}   & \multicolumn{1}{c}{77.8}   & \multicolumn{1}{c|}{83.9}   & \multicolumn{1}{c}{60.2}   & \multicolumn{1}{c|}{77.2}   & \multicolumn{1}{c}{71.4}   &  81.6  \\ \hline
\multicolumn{1}{|l|}{MASHQA (XLNet)}   & \multicolumn{1}{c}{69.4}   & \multicolumn{1}{c|}{71.3}   & \multicolumn{1}{c}{63.1}   & \multicolumn{1}{c|}{70.6}   & \multicolumn{1}{c}{67.1}   &  71  \\ \hline
\multicolumn{1}{|l|}{MSEQA (Roberta-Base)}   & \multicolumn{1}{c}{67.9}   & \multicolumn{1}{c|}{82.8}   & \multicolumn{1}{c}{56.5}   & \multicolumn{1}{c|}{77.8}   & \multicolumn{1}{c}{63.6}   &  81  \\ \hline
\multicolumn{1}{|l|}{MSEQA (Roberta-Large)}   & \multicolumn{1}{c}{80.2}   & \multicolumn{1}{c|}{85.5}   & \multicolumn{1}{c}{64.2}   & \multicolumn{1}{c|}{82.4}   & \multicolumn{1}{c}{74.4}   &  84.5  \\ \hline
\multicolumn{7}{|c|}{MASHQA}                                                                                                                                          \\ \hline
\multicolumn{1}{|l|}{MASHQA (XLNet)}   & \multicolumn{1}{c}{24.2}   & \multicolumn{1}{c|}{71.9}   & \multicolumn{1}{c}{21.7}   & \multicolumn{1}{c|}{54.2}   & \multicolumn{1}{c}{21.9}   &  57  \\ \hline
\multicolumn{1}{|l|}{MSEQA (Roberta-Base)}   & \multicolumn{1}{c}{25.3}   & \multicolumn{1}{c|}{74.3}   & \multicolumn{1}{c}{22.3}   & \multicolumn{1}{c|}{56.2}   & \multicolumn{1}{c}{22.8}   &  59  \\ \hline
\multicolumn{1}{|l|}{MSEQA (Roberta-Large)}   & \multicolumn{1}{c}{25.4}   & \multicolumn{1}{c|}{73.3}   & \multicolumn{1}{c}{22.7}   & \multicolumn{1}{c|}{56.6}   & \multicolumn{1}{c}{23.1}   & 59.2   \\ \hline
\end{tabular}
\caption{The performance of the baseline models along with our multispan question answer model, MSEQA, trained on combined set of all the datasets}
\end{table}

Next we study the individual component performances of MSEQA to identify any bottlenecks to the overall model capacity. In Table 3, we show the MSEQA capabilities to identify the question’s response as single span or multispan. Multispan classifier is very robust across the datasets in terms of identifying whether question has multiple span for its answer or just a single span. Table 4 displays  the answer-type predictor component’s isolated performance on the combination of the datasets. The high F1 score of 98.47 reflects its competence in classifying the answer as Yes, No or Unknown.

\begin{table}
   \label{T-appendix}
   \centering
    \begin{tabular}{ccclc}     
      \hline                   
    Dataset. & Precision & Recall & F1\\
      \hline
    DROP & 95.6 & 93.7  & 94.6 \\
    SMARTHOME & 97.6 & 96.7  & 97.4 \\
    MASHQA & 98.7 & 97.5  & 98.1 \\
      \hline
    \end{tabular}
    \caption{ MSEQA’s Multispan Predictor result }
   \end{table}

\begin{table}
   \label{T-appendix2}
   \centering
    \begin{tabular}{ccclc}     
      \hline                   
    Precision & Recall & F1\\
      \hline
    98.5 & 98.5  & 98.5 \\
      \hline
    \end{tabular}
    \caption{ Answer-type Predictor result }
   \end{table}
   
\begin{figure}[!h]
    \setlength\abovecaptionskip{0.2\baselineskip}
    \centering
    \includegraphics[width=.5\textwidth,clip=]{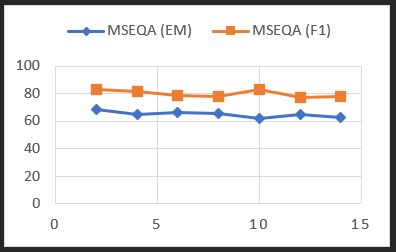}
    \caption{MSEQA result on across number of spans}
    \label{fig:nlp_fig4}
\end{figure}
We investigate the performance of MSEQA on increasing number of spans. We ran this experiment on SMARTHOME dataset as DROP and MASHQA did not have enough number of samples for more than 3 spans. From the Fig 4, we can observe that the MSEQA’s performance stays consistent across a wide range of spans reinforcing its robustness. We also study the performance of the models by changing the paragraph lengths. All the models show degradation in performance due to the attention span limitation of pretrained transformer-based encoders as investigated in [24]. Compared to the baselines, we can see the least performance drop with MSEQA showcasing its robustness with handling longer paragraph inputs.

\begin{figure}[!h]
    \setlength\abovecaptionskip{0.2\baselineskip}
    \centering
    \includegraphics[width=.4\textwidth,clip=]{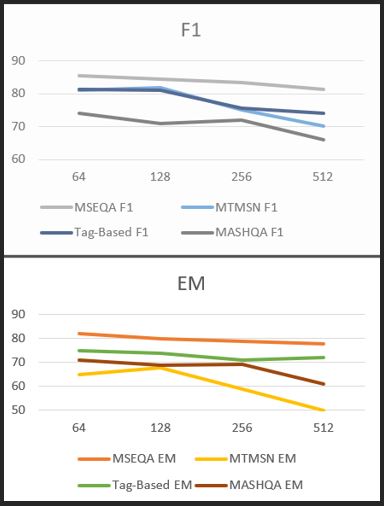}
    \caption{MSEQA result with different paragraph length}
    \label{fig:nlp_fig5}
\end{figure}

\section{Conclusion} 
      \label{S-features}      
We present a simple multispan architecture, MSEQA, for multi type question answering by classifying each sentence as a probable answer candidate. We show that when a combination of single/multi-span classifier with multispan tagging is used, the model provides robust answers for multi-span tasks without degrading its performance on single-span questions. As future work, we would like to further process selected sentences from the multi-span tagger and consolidate them into one fluent answer. We also plan to explore ways to put this question answering capability onto edge devices for various applications.

\begin{enumerate}
\item Xinyun Chen, Chen Liang, Adams Wei Yu, Denny Zhou, Dawn Song, and Quoc V Le. 2020. Neural symbolic reader: Scalable integration of distributed and symbolic representations for reading comprehension. In ICLR. 
\item Mutabazi, Emmanuel, Jianjun Ni, Guangyi Tang, and Weidong Cao. 2021. "A Review on Medical Textual Question Answering Systems Based on Deep Learning Approaches" Applied Sciences 11, no. 12: 5456. https://doi.org/10.3390/app11125456
\item https://broutonlab.com/blog/nlp-and-ai-for-ecommerce
\item Jacob Devlin, Ming-Wei Chang, Kenton Lee, and Kristina Toutanova. 2019. BERT: Pre-training of deep bidirectional transformers for language understanding. In NAACL.
\item Pranav Rajpurkar, Jian Zhang, Konstantin Lopyrev, and Percy Liang. 2016. SQuAD: 100,000+ questions for machine comprehension of text. In EMNLP. Lance Ramshaw and Mitch Marcus. 1995. Text chunking using transformation-based learning. In Third Workshop on Very Large Corpora. Erik F. Tjong Kim Sang. 2000. Transforming a chunker to a parser. In CLIN. 
\item Jambay Kinley and Raymond Lin. 2019. NABERT+: Improving numerical reasoning in reading comprehension. URL https: //github.com/raylin1000/drop-bert.
\item Dheeru Dua, Yizhong Wang, Pradeep Dasigi, Gabriel Stanovsky, Sameer Singh, and Matt Gardner. 2019. DROP: A reading comprehension benchmark requiring discrete reasoning over paragraphs. In Proc. of NAACL.
\item Ming Zhu, Aman Ahuja, Da-Cheng Juan, Wei Wei, and Chandan K. Reddy. 2020. Question Answering with Long Multiple-Span Answers. In Findings of the Association for Computational Linguistics: EMNLP 2020, pages 3840–3849, Online. Association for Computational Linguistics.
\item Zhilin Yang , PengQi, SaizhengZhang, YoshuaBengio, WilliamCohen, RuslanSalakhutdinov, and Christopher D.Manning. 2018. HotpotQA: A dataset for diverse, explainable multi-hop question answering. In Empirical Methods in Natural Language Processing (EMNLP), pages 2369–2380.
\item Yinhan Liu, Myle Ott, Naman Goyal, Jingfei Du, Mandar Joshi, Danqi Chen, Omer Levy, Mike Lewis, Luke Zettlemoyer, and Veselin Stoyanov. 2019. RoBERTa: A robustly optimized bert pretraining approach. arXiv preprint arXiv:1907.11692.
\item Zhiqing Sun, Hongkun Yu, Xiaodan Song, Renjie Liu, Yiming Yang, and Denny Zhou. 2020. MobileBERT: a Compact Task-Agnostic BERT for Resource-Limited Devices. In Proceedings of the 58th Annual Meeting of the Association for Computational Linguistics, pages 2158–2170, Online. Association for Computational Linguistics.
\item Zhenzhong Lan, Mingda Chen, Sebastian Goodman, Kevin Gimpel, Piyush Sharma, and Radu Soricut. ALBERT: A lite BERT for self-supervised learning of language representations. arXiv preprint arXiv:1909.11942, 2019.
\item Minghao Hu, Yuxing Peng, Zhen Huang, and Dongsheng Li. 2019. A multi-type multi-span network for reading comprehension that requires discrete reasoning. In Proceedings of EMNLP. 
\item Xuchen Yao, Benjamin Van Durme, Chris CallisonBurch, and Peter Clark. 2013. Answer extraction as sequence tagging with tree edit distance. In HLTNAACL.
\item Elad Segal, Avia Efrat, Mor Shoham, Amir Globerson, and Jonathan Berant. 2020. A simple and effective model for answering multi-span questions. In EMNLP
\item Zihang Dai, Zhilin Yang, Yiming Yang, William W Cohen, Jaime Carbonell, Quoc V Le, and Ruslan Salakhutdinov. Transformer-xl: Attentive language models beyond a fixed-length context. arXiv preprint arXiv:1901.02860, 2019.
\item Daniel Khashabi, Tushar Khot, Ashish Sabharwal, Oyvind Tafjord, Peter Clark, and Hannaneh Hajishirzi. 2020. Unifiedqa: Crossing format boundaries with a single qa system. In Findings of EMNLP.
\item Eunsol Choi , HeHe, MohitIyyer, MarkYatskar, Wen-tauYih, YejinChoi, PercyLiang, and LukeZettlemoyer. 2018. Quac: Question answering in context. In Proceedings of the 2018 Conference on Empirical Methods in Natural Language Processing, pages 2174–2184, Brussels.
\item Christopher Clark, Kenton Lee, Ming-Wei Chang, Tom Kwiatkowski, Michael Collins, and Kristina Toutanova. Boolq: Exploring the surprising difficulty of natural yes/no questions. arXiv preprint arXiv:1905.10044, 2019
\item Adam Paszke, Sam Gross, Francisco Massa, Adam Lerer, James Bradbury, Gregory Chanan, Trevor Killeen, Zeming Lin, Natalia Gimelshein, Luca Antiga, Alban Desmaison, Andreas Kopf, Edward Yang, Zachary DeVito, Martin Raison, Alykhan Tejani, Sasank Chilamkurthy, Benoit Steiner, Lu Fang, Junjie Bai, and Soumith Chintala. 2019. PyTorch: An imperative style, high-performance deep learning library. In H. Wallach, H. Larochelle, A. Beygelzimer, F. d'Alche-Buc, E. Fox, and R. Gar- ´ nett, editors, Advances in Neural Information Processing Systems 32, pages 8024–8035. Curran Associates, Inc. 
\item Thomas Wolf, Lysandre Debut, Victor Sanh, Julien Chaumond, Clement Delangue, Anthony Moi, Pierric Cistac, Tim Rault, R’emi Louf, Morgan Funtowicz, and Jamie Brew. 2019. HuggingFace’s transformers: State-of-the-art natural language processing. ArXiv, abs/1910.03771.
\item Diederik P Kingma and Jimmy Ba. Adam: A method for stochastic optimization. arXiv preprint arXiv:1412.6980, 2014
\item Rico Sennrich, Barry Haddow, and Alexandra Birch. 2016. Neural machine translation of rare words with subword units. In Association for Computational Linguistics (ACL), pages 1715–1725.
\item Yang Liu, Mirella Lapata. 2019. Text Summarization with Pretrained Encoders. Proceedings of Empirical Methods in Natural Language Processing and the 9th International Joint Conference on Natural Language Processing, pages 3730-3740.

\end{enumerate}

\end{document}